\pdfoutput=1

\documentclass[11pt]{article}

\usepackage{acl}
\usepackage{color}

\usepackage{times}
\usepackage{latexsym}
\usepackage{bm}
\usepackage{amsfonts}
\usepackage{amsmath}
\usepackage{multicol}
\usepackage{multirow}
\usepackage{booktabs}
\usepackage{graphicx}
\usepackage{array}
\usepackage{listings}
\usepackage{enumitem}

\usepackage[T1]{fontenc}

\usepackage[utf8]{inputenc}

\usepackage{microtype}

\title{
Listener Model for the \textit{PhotoBook} Referential Game \\
with CLIPScores as Implicit Reference Chain}

\author{
Shih-Lun Wu \and Yi-Hui Chou \and Liangze Li \\ 
\{\texttt{shihlunw, yihuic, liangzel}\}\texttt{@andrew.cmu.edu} \\
Language Technologies Institute, Carnegie Mellon University, Pittsburgh, PA, USA
}

\begin{document}
\maketitle
\begin{abstract}
PhotoBook is a collaborative dialogue game where two players receive private, partially-overlapping sets of images and resolve which images they have in common.
It presents machines with a great challenge to learn how people build common ground around multimodal context to communicate effectively.
Methods developed in the literature, however, cannot be deployed to real gameplay
since they only tackle some subtasks of the game,
and they require additional reference chains inputs, whose extraction process is imperfect.
Therefore, we propose a reference chain-free listener model
that directly addresses the game's predictive task, i.e., deciding whether an image is shared with partner.
Our DeBERTa-based listener model reads the full dialogue, and utilizes
CLIPScore features to assess utterance-image relevance.
We achieve $>$77\% accuracy on unseen sets of images/game themes, outperforming baseline by $>$17 points.
\end{abstract}

\section{Introduction}\label{sec:intro}
\begin{figure*}
    \centering
    \includegraphics[width=0.85\textwidth, trim={0mm, 25mm, 2mm, 20mm}, clip]{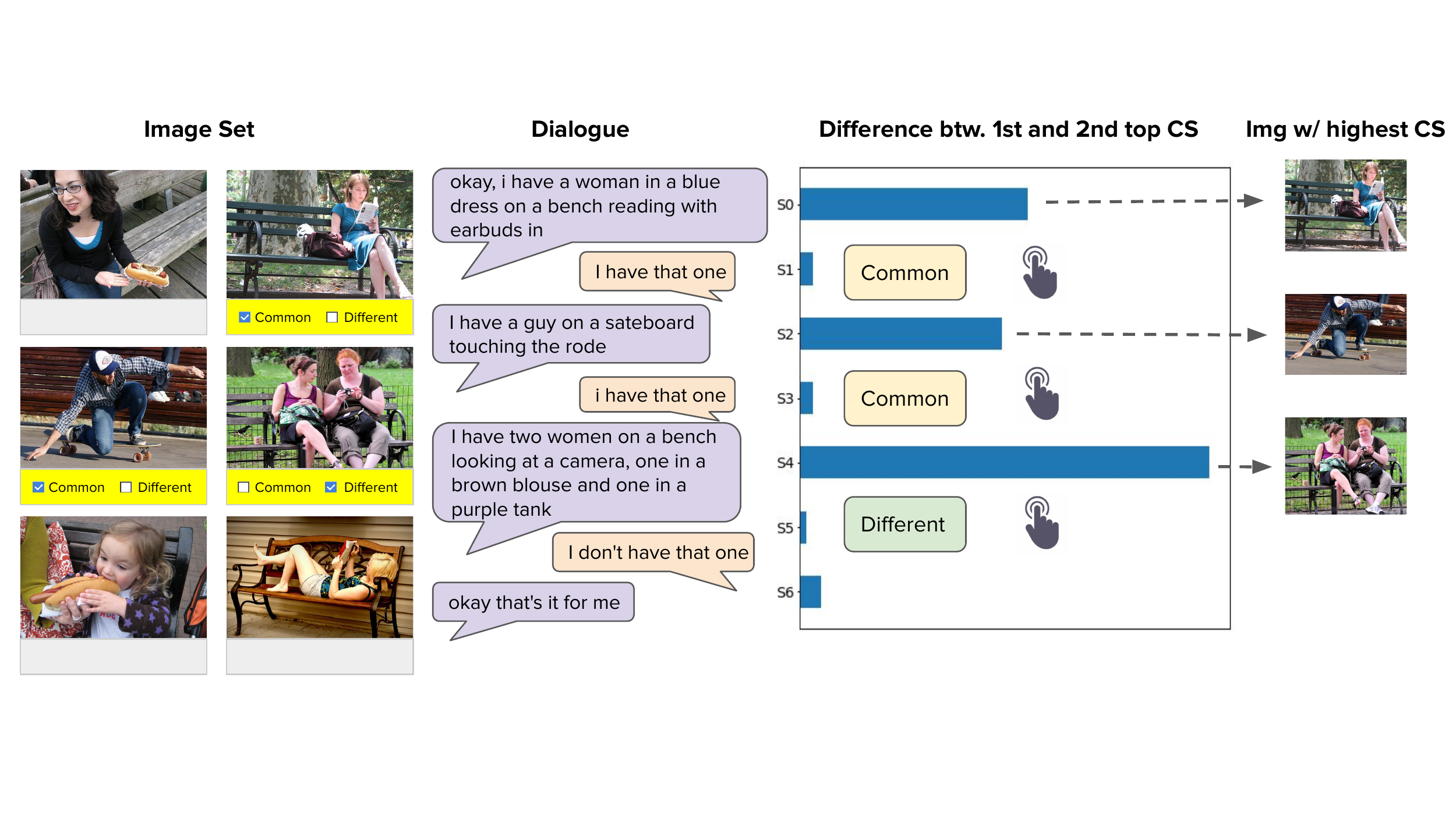}
    \vspace{-2mm}
    \caption{A round of PhotoBook game with dialogue, player marking actions, corresponding images, and CLIPScore (i.e., CS) difference between top and 2nd-top scoring images w.r.t.~the utterance. A player needs to figure out whether their partner has each of the 3 target (i.e., highlighted) images through text dialogue.}
    \label{fig:round_example}
\end{figure*}

PhotoBook \cite{haber2019photobook} is a collaborative dialogue game of two players.
In a game round, each player receives 6 images of an identical theme---the two largest objects in all images share the same categories, e.g., \textit{dog}, \textit{car}, etc.
The players have some of their images in common.
Their goal is to communicate through text dialogue, and individually mark 3 privately highlighted images
as either \textit{common} (i.e., shared with partner) or \textit{different}.
A full game lasts 5 rounds.
After each round, some of each player's images are replaced with different ones under the same theme.
Images may reappear in later rounds after being swapped out.
This game setup encourages building and leveraging common ground with multimodal contexts, which humans are known to do to facilitate conversation \cite{clark1986referring, brennan1996conceptual}.
Fig.~\ref{fig:round_example} displays an example of a PhotoBook game.\footnote{In this case, the game theme is \textit{person \& bench}.}



Models proposed in past works on the dataset \cite{haber2019photobook, takmaz2020refer} are unable to realistically play the game
due to several reasons:
(i) they only address subtasks in the game whose
time span is \textit{one utterance}, rendering it unnecessary for the models to keep track of the entire game's, or round's, progress; 
(ii) the models operate on additional input of \textit{reference chains}, i.e., past utterances referring to each image, whose (rule-based) extraction process is imperfect and hence complicates learning and evaluation;
and, (iii) utterances outside of reference chains, e.g., `\textit{I don't have that one}', may also be important pieces of information.


To address the drawbacks above,
we propose a full (i.e., able to play real games), reference chain-free listener model,
which accepts all dialogue utterances of a round\footnote{Though ideally, the model should process the entire game, i.e., 5 rounds, since formed consensus will be carried to subsequent rounds, doing so would lead to sequence lengths ($>$1K) longer than most pretrained Transformers have seen, necessitating an effective memory mechanism or extra adaptation efforts. Thus, we leave this setting for future endeavors.} and the 6 context images, and predicts whether the 3 target (highlighted) images are \textit{common/different}.
Our listener model is based on a pretrained DeBERTa Transformer \cite{he2020deberta}.
To incorporate visual context,
CLIPScores \cite{hessel2021clipscore} between each utterance and the 6 given images are infused with DeBERTa hidden states.
We employ CLIPScore as it offers strong prior knowledge about the relevance of an utterance to each of the 6 images,
which may serve as a soft, implicit version of reference chain used in previous studies.
Also, we chose DeBERTa since it is one of the top performers in the SuperGLUE benchmark \cite{sarlin2020superglue} which provides a reasonably-sized ($\sim$100M parameters) version to suit our purpose and computation resources.
We further devise a label construction scheme to create dense learning signals.
Our model scores a $>$77\% accuracy on the novel listener task and improves by $>$17\% (absolute) over the baseline adapted from \cite{takmaz2020refer}.
Our code is available at \url{github.com/slSeanWU/photobook-full-listener}.



\section{Related Work}\label{sec:related}
In typical collaborative dialogue tasks,
two agents (i.e., players) hold incomplete or partially overlapping information and communicate through text to reach a predefined goal.
The task-oriented setup enables simple evaluation for dialogue systems via task success rate, instead of resorting to costly human evaluation.
Tasks and datasets proposed in the literature focus either on set logic \cite{he2017learning}, image understanding \cite{de2017guesswhat, haber2019photobook}, or spatial reasoning \cite{udagawa2019natural}.
They challenge dialogue systems to process multiple modalities, discard irrelevant information, and build common ground.
Researchers have utilized graph neural networks \cite{he2017learning}, vision-and-language Transformers \cite{lu2019vilbert, tu2021learning}, and pragmatic utterance generation \cite{frank2012predicting, fried2021reference} to tackle the tasks.\footnote{Table \ref{tab:dataset} (in appendix) summarizes these tasks \& methods.}

To our knowledge, there has not been a system that fully addresses the PhotoBook task.
It may be particularly challenging due to the setup with multiple highly similar images and an unbounded set of information (e.g., scene, actions) the images may contain.
Previous PhotoBook works targeted two subtasks: \textit{reference resolution} \cite{haber2019photobook, takmaz2020refer} and \textit{referring utterance generation} \cite{takmaz2020refer}.
The former resolves which of the 6 context images an utterance is referring to, while the latter generates an informative utterance for a pre-selected image.
Proposed models take in extracted reference chains---whose rule-based extraction processes\footnote{Algorithmic details in Appendix \ref{sec:appendix}.} try to identify which utterances speak about each of the images.
To obtain such chains, \citet{haber2019photobook} broke the dialogue into segments using a set of heuristics based on player marking actions.
\citet{takmaz2020refer}, on the other hand, computed each utterance's BERTScore \cite{zhang2019bertscore} and METEOR \cite{banerjee2005meteor} respectively against ground-truth MSCOCO captions \cite{lin2014microsoft}, and VisualGenome attributes \cite{krishna2017visual} of each image to match (at most) one utterance per round to an image.

As for the reference resolution task, \citet{haber2019photobook} 
employed LSTM encoders.
One (query) encoder takes a current dialogue segment, while the other (i.e., context encoder) receives the 6 images' ResNet features, and the associated reference chain segments.\footnote{The 6 `images + ref.~chains' are processed separately.}
Dot products between query encoder output and 6 context encoder outputs are taken to predict the image the current segment refers to.
\citet{takmaz2020refer} largely kept the setup, but they used BERT \cite{devlin2019bert} embeddings
and contextualized utterances via weighted averaging instead of LSTMs.

\citet{takmaz2020refer} claimed an 85\% reference resolution accuracy, but they also reported an 86\% precision\footnote{evaluated on a human-labeled subset of 20 games} on reference chain extraction, making
it difficult to conclude whether prediction errors are due to model incompetence, or incorrect input data/labels. (We find that some parts of extracted reference chains either point to the wrong image or provide no information at all.\footnote{We rerun \cite{takmaz2020refer}'s experiment and show some of the problematic examples in Appendix \ref{sec:appendix} \& Table \ref{tab:wrong_pred}.})
Yet, we do agree that keeping track of which images have been referred to is vital for the game.
Therefore, we aim to build a full listener model that does not depend on explicit reference chains, but gathers such information from implicit hints given by an image-text matching model, i.e., CLIP \cite{radford2021learning}. 

\begin{figure}
    \centering
    \includegraphics[width=0.95\columnwidth, trim={90mm, 40mm, 50mm, 44mm}, clip]{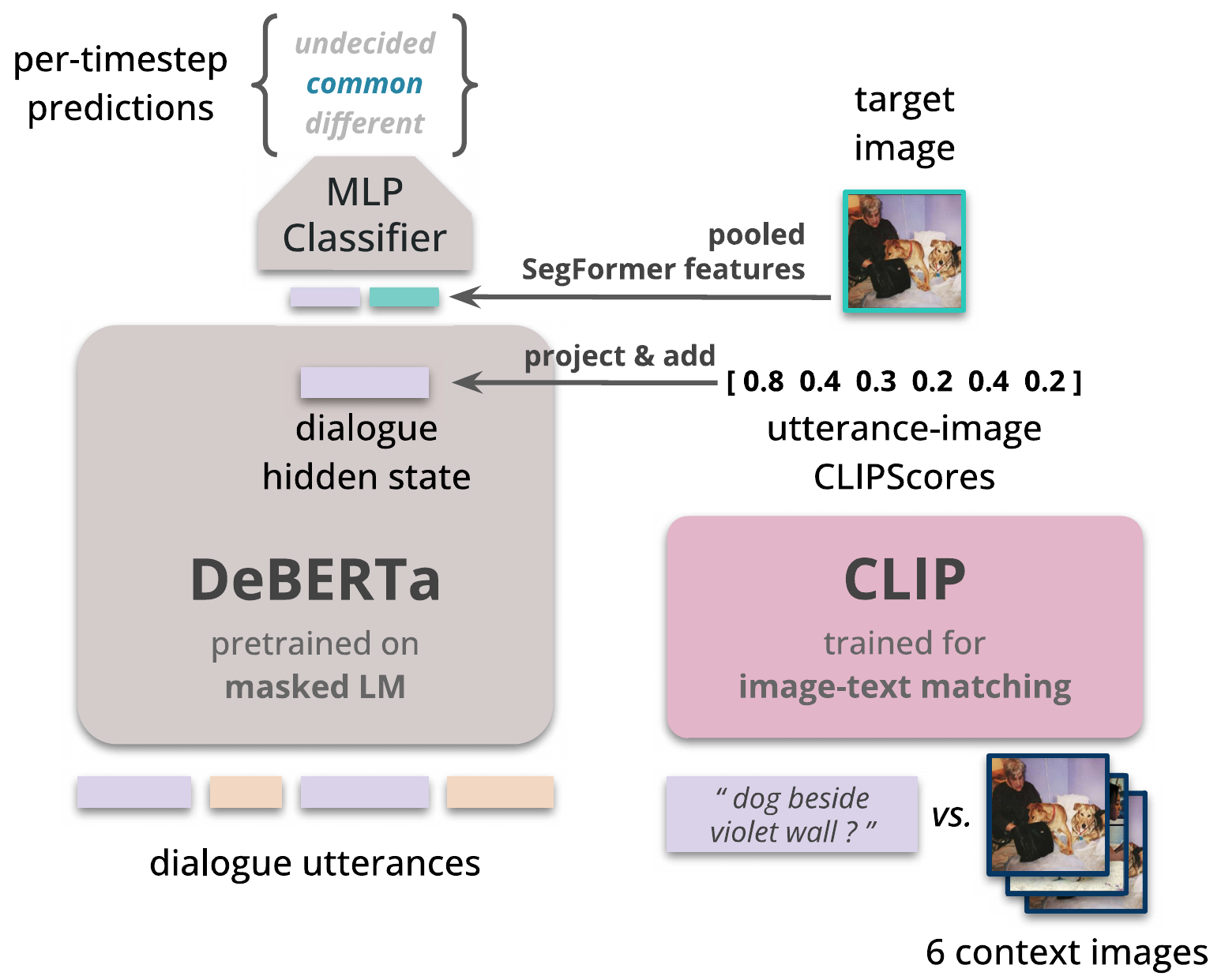}
    \vspace{-3mm}
    \caption{Overview of our listener model. A DeBERTa Transformer \cite{he2020deberta} encodes all utterances of a game round.
    Utterance-level CLIPScores \cite{hessel2021clipscore} w.r.t.~each image (i.e., an $\mathbb{R}^6$ vector) get projected and summed with hidden states of all timesteps corresponding to that utterance. Then, a 2-layer MLP takes in pooled SegFormer \cite{xie2021segformer} features of the target image ($\in \mathbb{R}^{512}$) and DeBERTa output to predict whether the image is \textit{common}, \textit{different}, or \textit{undecided} at every token timestep.}
    \label{fig:model}
    \vspace{-2mm}
\end{figure}

\section{Method}\label{sec:method}
\subsection{Functionality of CLIPScore}\label{subsec:clipscore}
Based on CLIP vision-and-language Transformer \cite{radford2021learning}, CLIPScore \cite{hessel2021clipscore} is a reference-free\footnote{i.e., does not take ground-truth text as input} metric to measure semantic image-text similarity.
On image captioning, \citet{hessel2021clipscore} showed that CLIPScore correlates better with human judgment than reference-dependent metrics like BERTScore \cite{zhang2019bertscore} and SPICE \cite{anderson2016spice}.


In our pilot study, we find that
the CLIPScore of an utterance-image pair is particularly high when the utterance describes the image
(see Fig.~\ref{fig:round_example} for example).
These score peaks thus form an \textit{implicit reference chain} for the dialogue, giving strong hints on whether the mentioned images are common/different when seen with subsequent partner feedback (e.g., `\textit{I have that one}').
Also, the reference chain extraction method in \cite{takmaz2020refer} achieves higher precision (86\%$\rightarrow$93\%) and recall (60\%$\rightarrow$66\%) when we simply replace its core scoring metrics\footnote{i.e., BERTScore \& METEOR. Details in Appendix \ref{sec:appendix}.} with CLIPScore.
The findings above show that CLIPScore captures well the utterance-image relationships in PhotoBook, and hence should be helpful to our listener model.

Computation-wise, reference chain extraction algorithms in the literature either rely on complex turn-level heuristics \cite{haber2019photobook}, or compute multiple external metrics (i.e., BERTScore and METEOR) \cite{takmaz2020refer}.
More importantly,
they have to wait until completion of a round to compute the chains.
Our utterance-level CLIPScores can be computed on the fly as utterances arrive, and are relatively time-efficient as they involve only one model (i.e., CLIP) and that batch computation may be used to increase throughput.

Modeling-wise,
reference chain extraction explicitly selects which utterances the listener model should see, so when it is wrong, the model either sees something irrelevant, or misses important utterances.
On the other hand, utterance-level CLIPScores resemble using a highlighter to mark crucial dialogue parts for the model.
Even when CLIPScores are sometimes inaccurate, the model could still access the full dialogue to help its decisions.

\subsection{The Full Listener Model}\label{subsec:model}

\subsubsection{Inputs}
An overview of our listener model is depicted in Fig.~\ref{fig:model}.
Our model operates on three types of input features, which collectively represent a game round from one of the players' perspective:
\vspace{-1mm}
\begin{align}
    \text{Dialogue tokens:}  \; \mathcal{X} &= \{\bm{x}_k \in \mathcal{W}^{|\mathcal{T}_k|}\}_{k=1}^K \\
    \text{CLIPScores:} \; \mathcal{C} &= \{\bm{c}_k \in \mathbb{R}^{6}\}_{k=1}^K \\
    \text{Image features:}  \; \mathcal{V} &= \{\bm{v}_{j} \in \mathbb{R}^{512}\}_{j=1}^{6} \label{eqn:pooled-vis-inp}
\end{align}
\noindent We use $k$, $j$ to index utterances and images respectively. $\mathcal{W}$ is the text token vocabulary, and $\mathcal{T}_k = \{t_{k, \text{start}},\dots,t_{k, \text{end}}\}$ is the corresponding token timesteps for the $k^{\text{th}}$ utterance.
To the start of each utterance, we prepend either a \textsf{[CLS]} or \textsf{[SEP]} token to distinguish whether it comes from the player itself or the partner.
All utterances are concatenated to form one text input sequence to our model.\footnote{Average text length (i.e., $\sum_k|\mathcal{T}_k|$) is about 120 tokens.}
CLIPScore vectors ($\bm{c}_k$'s) are computed 
in a per-utterance manner, i.e.,
between one utterance and each of the 6 images.
Images are represented by the pooled\footnote{Pooling of the 16$\times$16 SegFormer patch features per image into one involves 2d-conv.~downsampling other than taking the mean, as we also attempt fusing visual context by cross-attending to patch features. More details in Appendix \ref{sec:appendix-visattn}.} features from SegFormer \cite{xie2021segformer}.
It is trained on semantic image segmentation \cite{zhou2017scene}, and hence should encode crucial visual information for the game, i.e., objects in the scene and their spatial relationships.

\subsubsection{Labels and Output}\label{subsec:labels}
Rather than training the model to predict just once after seeing the entire dialogue,
we construct labels for \textit{all} timesteps, forming a label sequence $\bm{y}_j \in \mathcal{L}^T$, where $T = \sum_k|\mathcal{T}_k|$, for each target image, where $\mathcal{L}$ is the label set.
As there are only 3 target images out of the 6, we also only have 3 such label sequences ($\bm{y}_j$'s) for a training instance.
At each timestep $t$, the label of a target image, $y_{j,t} \in \mathcal{L}$, is one of $\{\textit{undecided}, \textit{common}, \textit{different}\}$.
It always starts as \textit{undecided}, changes to \textit{common} or \textit{different} at the moment of player marking action, and remains there for the rest of the dialogue.
Our model's output for a (target) image $j$ at timestep $t$ is hence a distribution $\bm{\hat{y}}_{j, t} \in \mathbb{R}^3$,
which is a temporary belief about that image.
Also, we apply causal masking on DeBERTa self-attention.
Such a labeling and masking scheme creates dense learning signals---our model must judge an image at every timestep based on growing dialogue context.

\subsubsection{Model Components}
The backbone of our model is a pretrained base DeBERTa \cite{he2020deberta}, which takes in concatenated utterances $\mathcal{X} = \{\bm{x}_k \in \mathcal{W}^{|\mathcal{T}_k|}\}_{k=1}^K = \{x_t \in \mathcal{W}\}_{t=1}^T$,
and contextualizes them into hidden states:
\begin{equation}
    \mathcal{H}^{(l)} = \{\bm{h}^{(l)}_t \in \mathbb{R}^d\}_{t=1}^T \, , \; \; l \in \{1,\dots,L\} \, ,
\end{equation}
where $d$ ($=\,$768) is DeBERTa's hidden size, and $l$ is layer index (\# layers $L =\,$12).
We do not adopt vision-and-language Transformers \cite{lu2019vilbert, wang2022ofa} for they are pretrained on `single image-short text' pairs, which mismatches our scenario.
Following \citet{wu2021musemorphose}'s recommendation on feeding time-varying conditions to Transformers, utterance-level CLIPScores (i.e., $\mathcal{C}$) are projected and summed with DeBERTa hidden states at \textit{all} layers:\footnote{Additional experiments in Appendix \ref{sec:appendix-layers} shows that feeding CLIPScore to fewer layers harms the performance.}
\begin{equation}
    \mathcal{H}^{(l)} \leftarrow  \; \{\mathcal{H}_{\mathcal{T}_k}^{(l)} = \bm{h}^{(l)}_{t \in \mathcal{T}_k} + \bm{W}_\text{proj} \, \bm{c}_k\, \}_{k=1}^K \, ,
\end{equation}
where $\bm{W}_\text{proj} \in \mathbb{R}^{d\times6}$ is a learnable matrix.

To make predictions, we place a 2-layer MLP (with GELU activation) on top of DeBERTa.
It takes in the concatenation of the pooled target image features and the last-layer DeBERTa hidden state, and produces a distribution over the label set $\mathcal{L} = \{\textit{undecided}, \textit{common}, \textit{different}\}$:
\begin{equation}
    \bm{\hat{y}}_{j, t} = \mathrm{MLP}_{\mathbb{R}^{512 +d}\rightarrow\mathbb{R}^3}([\bm{{v}}_j; \bm{h}^{(L)}_t]) \, . \label{eqn:mlp}
\end{equation}
We add learnable positional embeddings to $\bm{{v}}_j$'s to make our model aware of the target image's index.

\begin{table}
\small
\centering
\begin{tabular}{l | c c }
\toprule
 & valid & test \\
\midrule
\textit{Random guess} & 50.0 \; \; \; \, & 50.0 \; \; \; \, \\
\hline
Modified \cite{takmaz2020refer} & 64.2\scriptsize{$\:\pm$\,1.7} & 59.0\scriptsize{$\:\pm$\,0.7} \\
\; w/ CLIPScore ref chains & 65.0\scriptsize{$\:\pm$\,1.4} & 59.7\scriptsize{$\:\pm$\,0.8} \\
\hline
\textbf{Ours} & \textbf{84.8}\scriptsize{$\:\pm$\,1.3} & \textbf{77.3}\scriptsize{$\:\pm$\,0.3} \\
\; a.\textbf{ $+$\,VisAttn} & 75.0\scriptsize{$\:\pm$\,0.6} & 69.8\scriptsize{$\:\pm$\,3.3} \\
\; b.\textbf{ $-$\,CLIPScore} & 70.7\scriptsize{$\:\pm$\,1.1} & 64.8\scriptsize{$\:\pm$\,1.5} \\
\; c.\textbf{ $-$\,CLIPScore $+$\,VisAttn} & 69.8\scriptsize{$\:\pm$\,1.1} & 64.9\scriptsize{$\:\pm$\,0.4} \\
\; d.\textbf{ $-$\,Dense learning signals} & 59.4\scriptsize{$\:\pm$\,1.8} & 55.9\scriptsize{$\:\pm$\,0.9} \\
\hline
\textit{Human} & 95.0 \; \; \; \, & 94.5 \; \; \; \, \\
\bottomrule
\end{tabular}
\vspace{-2mm}
\caption{Listener model accuracy (\%) of baselines and our model (full \& ablated versions). StDev of 3 runs with fixed seeds shown after $\pm$.
Pairwise bootstrap tests corroborate ($p <\,$ .001) that our full model outperforms all baselines and ablated versions.
\textit{Human} is the accuracy annotators achieved during dataset creation.
(\textbf{VisAttn}: cross-attention to patch features of 6 context images.)}\label{tab:result}
\vspace{-2mm}
\end{table}

\section{Experiments and Results}\label{sec:results}
Our listener model is trained with the maximum likelihood estimation (MLE) loss function:
\begin{equation}
    \mathbb{E}_{(\mathcal{X}, \mathcal{C}, \mathcal{V}, \mathcal{Y}) \in \mathcal{D}_\text{train}} \sum_{j, t} - \log p_{\bm{\hat{y}}_{j, t}} (y_{j, t} \, | \, \mathcal{X}, \mathcal{C}, \bm{v}_j) , \label{eqn:objective}
\end{equation}
where $\mathcal{D}_\text{train}$ is the training split, and $\mathcal{Y}$ is the set of label sequences associated with a data instance.
The same images/themes are guaranteed not to appear in multiple dataset splits.
We refer readers to Appendix \ref{sec:appendix-impl} for more implementation and training details.
Evaluation metric adopted here is accuracy measured at the end of dialogue, i.e., at evaluation, we ignore temporary beliefs in the chat.
To set a baseline, we modify the reference resolution model in \cite{takmaz2020refer} to suit our listener task.\footnote{Modification details are in Appendix \ref{appendix:baseline}.}

Table \ref{tab:result} lists the evaluation results.
Our method outperforms baseline by 17$\sim$20 percentage points, closing the gap to human performance by more than half.
Examining the ablations, we can observe that both removing CLIPScore inputs and dense learning signals (i.e., having labels at all timesteps, see Sec.~\ref{subsec:labels}) cause serious accuracy degradation, indicating their essentiality in our model, and that a pretrained Transformer does not trivially beat a fully MLP-based baseline.
Besides, though adding cross-attention to image features\footnote{Cross-attention mechanism explained in Appendix \ref{sec:appendix-visattn}.} (i.e., ablations a.~\& c.)~seems to be a more intuitive way to involve visual context, it leads to more severe overfitting\footnote{Likely due to limited dataset size and configuration. More analysis and exploration can be found in Appendix \ref{appendix:overfit}.} and hence does not help in our case.
We provide more detailed observations on our best-performing model's behavior and outputs in Appendix \ref{sec:predmistake}.

 \section{Conclusions and Future Work}\label{sec:conclusion}
In this paper, we first discussed why it is difficult to deploy existing reference chain-dependent PhotoBook models to real gameplay,
and demonstrated that CLIPScore's image-text matching capability may provide implicit reference chains to the task.
We then developed a novel listener model that 
is reference chain-free,
and able to realistically play the game given text dialogue and the set of context images, just as what human players see.
The model is built on a DeBERTa Transformer backbone,
and brings in visual context by infusing utterance-level CLIPScores with its hidden states.
On the newly proposed full listener task, i.e., predicting whether an image is shared with partner,
our model achieves 77$\sim$84\% accuracy on unseen sets of images, surpassing baseline \cite{takmaz2020refer} by over 17 points.
Ablation studies also showed that feeding CLIPScores and imposing dense learning signals are both indispensable to our model's success.

Future studies may leverage parameter-efficient transfer learning \cite{he2021towards, houlsby2019parameter, hu2021lora, perez2018film}
to cope with image data scarcity of PhotoBook (and potentially other datasets and tasks).
It is also interesting to develop a speaker model that uses temporary beliefs from our listener model and takes pragmatics \cite{frank2012predicting, fried2021reference} into account to generate informative responses.
Pairing such a model with our listener model may complete the collaborative dialogue task end-to-end.

\section{Limitations}\label{sec:limits}
The PhotoBook dataset has a very limited number of images (i.e., 360) and image combinations (i.e., 5 per game theme), which may lead to undesirable overfitting behavior as we discuss in Appendix \ref{appendix:overfit}.
Also, since our model depends heavily on CLIP \cite{radford2021learning}, it is likely to inherit CLIP's biases and weaknesses.
For example, \citet{radford2021learning} mentioned that CLIP fails to perform well on abstract or more complex tasks, such as counting or understanding spatial relationships between objects.
Finally, whether our listener model can be easily applied/adapted to productive real-world tasks (e.g., automated customer service with image inputs) requires further exploration.

\section*{Acknowledgements}

We would like to express our utmost thanks to Dr.~Daniel Fried, Emmy Liu and Dr.~Graham Neubig for their guidance and insightful suggestions.
We also appreciate the valuable feedback from the reviewers and the area chair.


\bibliography{anthology,custom}
\bibliographystyle{acl_natbib}


\noindent \textbf{\large{Appendices}}
\appendix
\section{Details on Model Implementation and Training}\label{sec:appendix-impl}
Our listener model's implementation is based on HuggingFace's DeBERTa module.\footnote{\url{github.com/huggingface/transformers/blob/main/src/transformers/models/deberta/modeling_deberta.py}}
The 16$\times$16 (512-dimensional) patch features  for each context image are extracted from last encoder layer of the publically released SegFormer-b4 model\footnote{\url{huggingface.co/nvidia/segformer-b4-finetuned-ade-512-512}} trained on ADE20k \cite{zhou2017scene} semantic image segmentation dataset.
CLIPScores between utterances and images are computed using the official repository\footnote{\url{github.com/jmhessel/clipscore}} which employs Vision Transformer-base (ViT-B/32) \cite{dosovitskiy2020image} as the image encoder.
Our listener model adds $\sim$1M trainable parameters to the 12-layer base DeBERTa backbone, which originally has 100M parameters.

We split our dataset to train/validation/test with a 70/10/20 ratio and make sure that a theme (i.e., categories of the 2 largest objects appearing in all 6 context images in a game round), and hence any image, does not lie across multiple splits.
Since a game round has 2 perspectives (i.e., players), it also spawns 2 instances.
Rounds in which players make mistakes, or mark images before the first utterance, are filtered out.
We finally obtain 13.7K/1.8K/3.7K instances for each of the splits respectively.

We train the model for 100 epochs and early stop on validation accuracy with 10 epochs of patience.
AdamW \cite{loshchilov2018decoupled} optimizer with $10^{-3}$ weight decay is used. We warm up the learning rate linearly for 500 steps to $2\times10^{-5}$, and then linearly decay it to $0$ for the rest of the training.
Batch size is set to 16.
Training takes around 8 hours to complete on an NVIDIA A100 GPU with 40G memory.
For fair comparison across model settings and baselines, we randomly draw 3 seeds and run training on all settings/baselines with them.

\begin{table*}
\scriptsize
\setlength{\tabcolsep}{0.45em}
\centering
\begin{tabular}{l | c c c c c}
\toprule
 & Dataset size & Inputs & Tgt.~resolution & SoTA E2E performance & SoTA techniques \\
\midrule
\textit{MutualFriends} \cite{he2017learning} & 11K dialogues & Text (tabular) & Bilateral & 96\% \cite{he2017learning} & GNN, LSTM \\
\textit{GuessWhat?!} \cite{de2017guesswhat} & 150K dialogs, 66K imgs & Text \& image & Unilateral & 63\% \cite{tu2021learning} & ViLBERT \\
\textit{OneCommon} \cite{udagawa2019natural} & 5K dialogues & Text \& dots on plane & Bilateral & 76\% \cite{fried2021reference} & LSTM, CRF, RSA \\
\textit{PhotoBook} \cite{haber2019photobook} & 12.5k dialogs, 360 imgs & Text \& 6 images & Bilateral & No complete system yet & ResNet, LSTM \\
\bottomrule
\end{tabular}
\vspace{-2mm}
\caption{Some datasets for collaborative dialogue tasks.
Bilateral (or unilateral) `Tgt.~resolution' means whether it requires both (or just one) players to figure out the entities/objects they should focus on. 
(Performance is measured by end-to-end task success.)}\label{tab:dataset}
\end{table*}

\section{Details on the Attempt to Infuse Visual Features with Cross Attention}\label{sec:appendix-visattn}
In addition to fusing CLIPScores into DeBERTa self-attention, we also attempt cross-attending DeBERTa hidden states to the 6 context images' SegFormer features to incorporate visual information.

We denote the SegFormer patch features by:
\begin{equation}
    \mathcal{V}^{\mathrm{(pt)}} = \{\bm{v}_{j, p}^\mathrm{(pt)} \in \mathbb{R}^{512}\}_{j=1, \; p=1}^{6, \; \; \; \; \; 16\times16} \, ,
\end{equation}
where $j, p$ respectively indexes images and patches.
All image features (16$\times$16$\times$6$\,=\,$1536 vectors) are concatenated into one long sequence for the DeBERTa hidden states (with text \& CLIPScore information) to cross-attend to.
As a sequence with length over 1.5K would lead to large memory footprint for attention operations, we downsample the patch features (to 8$\times$8$\times$6 $\,=\,$ 384 vectors) through strided 2D group convolution before feeding them to cross-attention, i.e., 
\begin{align}\label{eqn:patch-attn-first}
    \mathcal{\dot{V}}^{{\mathrm{(pt)}}} = \; &\mathrm{StridedGroupConv2D}(\mathcal{V}^{{\mathrm{(pt)}}})\\ 
    \mathcal{H}^{(l)} \leftarrow \; &\mathrm{Attention}(\mathcal{H}^{(l)}, \mathcal{\dot{V}^{\mathrm{(pt)}}}, \mathcal{\dot{V}^{\mathrm{(pt)}}}) \, ,\label{eqn:patch-attn-last}
\end{align}
where $\mathcal{H}^{(l)}$ is the $l^{\text{th}}$-layer DeBERTa hidden states.
The patch features in $\mathcal{\dot{V}^{\mathrm{(pt)}}}$ are further mean-pooled to form inputs (for target images), i.e., $\mathcal{V}$, to our final MLP classifier (please check Eqn.~\ref{eqn:pooled-vis-inp} \& \ref{eqn:mlp}, too):
\begin{equation}
    \mathcal{V} = \{ \bm{v}_j \}_{j=1}^{6} = \{ \: \mathrm{MeanPool}(\{\bm{\dot{v}}^{\mathrm{(pt)}}_{j, p}\}_{p=1}^{8\times8}) \: \}_{j=1}^{6} \\
\end{equation}

In the model settings whose performance is reported in Table \ref{tab:result} (i.e., ablations a.~\& c.), we place two such cross-attention layers with tied weights before all DeBERTa self-attention layers
to give the model more chances to digest and reason with visual inputs.
Doing so introduces 8M new trainable parameters (cf.~$\sim$1M for our best model). 
We also try to place these cross-attention layers later in the model in unreported experiments.
However, when using visual cross-attention, our listener model always suffers more from overfitting---lower training loss but worse evaluation accuracy.


\section{Adapting \citet{takmaz2020refer}'s Model for Our Listener Task}\label{appendix:baseline}
The reference resolution model in \cite{takmaz2020refer} contains two components: query encoder and context encoder:
\begin{itemize}[topsep=4pt, itemsep=4pt, parsep=0pt, leftmargin=*]
    \item Query encoder: takes in BERT embeddings of a \textit{current utterance} and the concatenation of 6 context images' ResNet features, and outputs one representation through learnable weighted averaging (across utterance timesteps).
    \item Context encoder: encodes each of the 6 images and the associated \textit{reference chain} (i.e., past utterances referring to that image) separately.
    The average of each reference chain utterance's BERT embeddings gets summed with that image's ResNet features to form the context representation for that image.
\end{itemize}
The model is based on fully-connected layers entirely.
Finally, dot products between the query representation and 6 context representations are taken, and the $\mathrm{arg\,max}$ is deemed the referent image of the current utterance.

To adapt their model to our full listener task, we feed to the query encoder BERT embeddings of the \textit{whole round of dialogue} and ResNet features of the \textit{target image} instead.
We \textit{mean-pool} the 6 context encoder representations, concatenate this pooled representation with the query representation, and apply a GELU-activated 2-layer MLP (similar to our model's) on top of the concatenated representations to predict whether the target image is \textit{common} or \textit{different}.
This modified baseline model can hence be trained using an objective similar to our model's (i.e., Eqn.~\ref{eqn:objective}).
Note that there is no dense learning signal for this adapted baseline, as the representation from query encoder is already pooled across timesteps.

\section{Experiments on CLIPScore Injection Layers}\label{sec:appendix-layers}
\begin{table}[h]
\centering
\begin{tabular}{l | c c }
\toprule
 Layers fed & valid & test \\
\midrule
\textbf{[emb]} & 72.4\footnotesize{$\:\pm$\,0.7} & 66.3\footnotesize{$\:\pm$\,0.5} \\
\textbf{[emb, 1st]} & 78.7\footnotesize{$\:\pm$\,1.4} & 71.9\footnotesize{$\:\pm$\,1.6} \\
\textbf{[emb, 1st$\sim$5th]} & 82.2\footnotesize{$\:\pm$\,1.0} & 76.5\footnotesize{$\:\pm$\,1.1} \\
\textbf{[4th$\sim$9th]} & 82.7\footnotesize{$\:\pm$\,0.7} & 76.1\footnotesize{$\:\pm$\,0.6} \\
\textbf{[7th$\sim$12th]} & 83.0\footnotesize{$\:\pm$\,0.6} & 75.9\footnotesize{$\:\pm$\,0.6} \\
\textbf{All layers} & \textbf{84.8}\footnotesize{$\:\pm$\,1.3} & \textbf{77.3}\footnotesize{$\:\pm$\,0.3} \\
\hline
w/o CLIPScores & 70.7\footnotesize{$\:\pm$\,1.1} & 64.8\footnotesize{$\:\pm$\,1.5} \\
\textit{Human} & 95.0 \; \; \; \, & 94.5 \; \; \; \, \\
\bottomrule
\end{tabular}
\caption{Accuracy (\%) of our listener model with CLIPScores fed to various layers. StDev of 3 runs with specific random seeds shown after $\pm$.}\label{tab:cslayer}
\end{table}

\noindent \citet{wu2021musemorphose} maintained that feeding time-varying conditions to Transformers more times over the attention layers enhances the conditions' influence, and hence improves performance.
Therefore, we choose to infuse CLIPScores with DeBERTa at all attention layers by default.
Table~\ref{tab:cslayer} shows the performance when we inject CLIPScores to fewer layers.
As expected, the more layers CLIPScores are fed to, the better the performance (6 layers $>$ 2 layers $>$ 1 layer, all with $p <\,$ .01).
Yet, infusing at earlier or later layers (3$^\text{rd}\sim$5$^\text{th}$ columns in Table \ref{tab:cslayer}) does not make a meaningful difference. 


\section{Experiments on Overfitting Behavior}\label{appendix:overfit}

\begin{table}[h]
\small
\centering
\setlength{\tabcolsep}{0.45em}
\begin{tabular}{l | c c c }
\toprule
 & val \textbf{(I)} & val \textbf{(P)} & test \textbf{(I/P)} \\
\midrule
\textbf{Full model} & 63.7 & 97.4 & 71.2 / 76.6 \\
\; b.\textbf{ $-$\,CLIPSc} & 58.6 & 91.7 & 63.8 / 63.6 \\
\; c.\textbf{ $-$\,CLIPSc $+$\,VisAttn} & 57.4 & 99.1 & 63.9 / 57.2 \\ 
\bottomrule
\end{tabular}
\caption{Accuracy (\%) with repartitioned train/val sets. Test sets \textbf{(I)/(P)} are identical and are the same as the one used in Tables \ref{tab:result} \& \ref{tab:cslayer}. They are meant to report test accuracy under \textbf{(I)/(P)} partitioning. All results are from the same random seed.}\label{tab:overfit}
\end{table}

\noindent \citet{haber2019photobook} stated that to collect a sufficient number of reference chains for each game theme, only 5 unique combinations (of two sets of 6 images) were picked and shown to the players.\footnote{in the 5 rounds of a game with randomized order}
This number is drastically smaller than the total \# of possible combinations.
(Suppose we want the players to have 2$\sim$4 images in common, then there would be $\binom{12}{6}\binom{6}{2}\binom{10}{4} + \binom{12}{6}\binom{6}{3}\binom{9}{3} + \binom{12}{6}\binom{6}{4}\binom{8}{2} \approx \,$ 4.85M combinations.)
Also, we observe that models with full access to image features (i.e., those with visual cross-attention) exhibit worse overfitting.
Hence, we suspect that our model overfits to specific image combinations, i.e., memorizing the labels from them.
To test this hypothesis out, we repartition our train \& validation sets such that a game theme appears in both sets, but in two different ways:
\begin{itemize}[topsep=4pt, itemsep=4pt, parsep=0pt, leftmargin=*]
    \item train/val \textbf{(I)}: val set has \textbf{unseen} image combinations, but \textbf{seen} pairs of players
    \item train/val \textbf{(P)}: val set has \textbf{unseen} pairs of players, but \textbf{seen} image combinations
\end{itemize}
The test set is left unchanged.
We train the models for 50 epochs without early stopping here.

Performance resulting from these repartitions is shown in Table \ref{tab:overfit}.
The numbers support our hypothesis in general.
Across different settings, our model does almost perfectly when an image combination (and hence the correct common/different answers) is seen during training (i.e., val \textbf{(P)}), and fails when being presented with a new image combination of a seen game theme.
As anticipated, the accuracy gap is the worst when visual cross-attention is added.
Moreover, it is worth mentioning that our models perform even worse on `seen images, unseen image combination' (i.e., val \textbf{(I)}) than on `unseen images' (i.e., test set).
Therefore, we conjecture that, with such a limited number of images and image combinations, it becomes trivial for deep models to exploit the (prescribed) relationships between inputs and labels, 
hindering the desirable learning goal---knowing the differences across similar images, and identifying crucial ones for the predictive task with the help of dialogue.
This is a major limitation of the PhotoBook dataset.

\section{The (Imperfect) Reference Chain Extraction Process}\label{sec:appendix}
Previous works on reference resolution \cite{haber2019photobook, takmaz2020refer} require extracted reference chains for training and evaluation.
We rerun experiments for the reference resolution model in \cite{takmaz2020refer} and get an 85\% accuracy (on reference resolution, not our full listener task), which is similar to the reported number.
Upon examining the predictions, we find that 9 out of 10 wrong predictions (w.r.t.~extracted labels) with the highest confidence are caused by problematic input data/labels resulting from reference chain extraction.
These cases are either due to mislabeled ground truth (while the model actually makes a reasonable prediction), low-quality utterances that provide vague or irrelevant information, reference chains not consistently pointing to one image, or a mix of all the above.
Table \ref{tab:wrong_pred} presents some examples.

\section{Further Observations on Our Listener Model Behavior and Outputs}\label{sec:predmistake}

First, we are interested in how characteristics of those $\mathbb{R}^6$ CLIPScore vectors might influence our listener model's decisions.
As mentioned in Sec.~\ref{subsec:clipscore}, an image tends to get a much higher CLIPScore when being spoken about by the utterance.
Therefore, we look at the 3 CLIPScore vectors per round with the largest difference between highest and 2$^\text{nd}$-highest CLIPScore values.\footnote{A player has to deal with 3 images per round, and we observe that in most cases, there is one utterance talking specifically about each image.}
We then group rounds (in test set) according to whether the model predicts all 3 target images correctly as \textit{common} or \textit{different}.\footnote{The model gets 1.7K out of 3.7K samples entirely correctly, while the rest have 1$\sim$3 wrong predictions.}
For the \textit{all-correct} cases, the difference between the top two values in the CLIPScore vectors (3 per round, as said above) has a mean$\,=\,$0.112 (std$\,=\,$0.063), whereas in the cases where the model makes one or more mistakes, the mean is 0.101 (std$\,=\,$0.062).
Unpaired t-test indicates a significant difference ($p <\,$ .001) between the pair of statistics.
This suggests a possibility that our model works better when CLIPScores contrast different images more clearly.

Next, we inspect the cases where our model predicts all 3 target images incorrectly.
Out of 111 such rounds, 72 are concentrated in two themes, i.e., \textit{cup \& dining table}, and \textit{car \& motorcycle}.
Images in the two themes are usually more difficult to be told apart.
Human players also score a lower 94.1\% accuracy on either of the two themes, compared to the 95.3\% overall, and 94.5\% over the test set.
Table \ref{tbl:look_at_all_wrongs} displays two examples of such \textit{all-wrong} rounds (respectively from \textit{cup \& dining table} and \textit{car \& motorcycle} game themes).
In the first example, target images 1 and 2 are highly similar such that player used `sandwhich' and `mug' to describe both of them.
In the second example, apart from similar images, multiple questions were thrown at the same time and answered as many as 4 utterances later.
Typos (e.g., \textit{sandwhich}, \textit{vlack}) and automatically filtered words (e.g., \textit{m**fin}) may also confuse the model.
However, we note that with so many inputs (i.e., text, CLIPScores, pooled target image feature) to our listener model, it is not straightforward to figure out the actual causes of wrong predictions.


\begin{table*}
    \includegraphics[width=\textwidth, trim={5mm, 200mm, 5mm, 5mm}, clip]{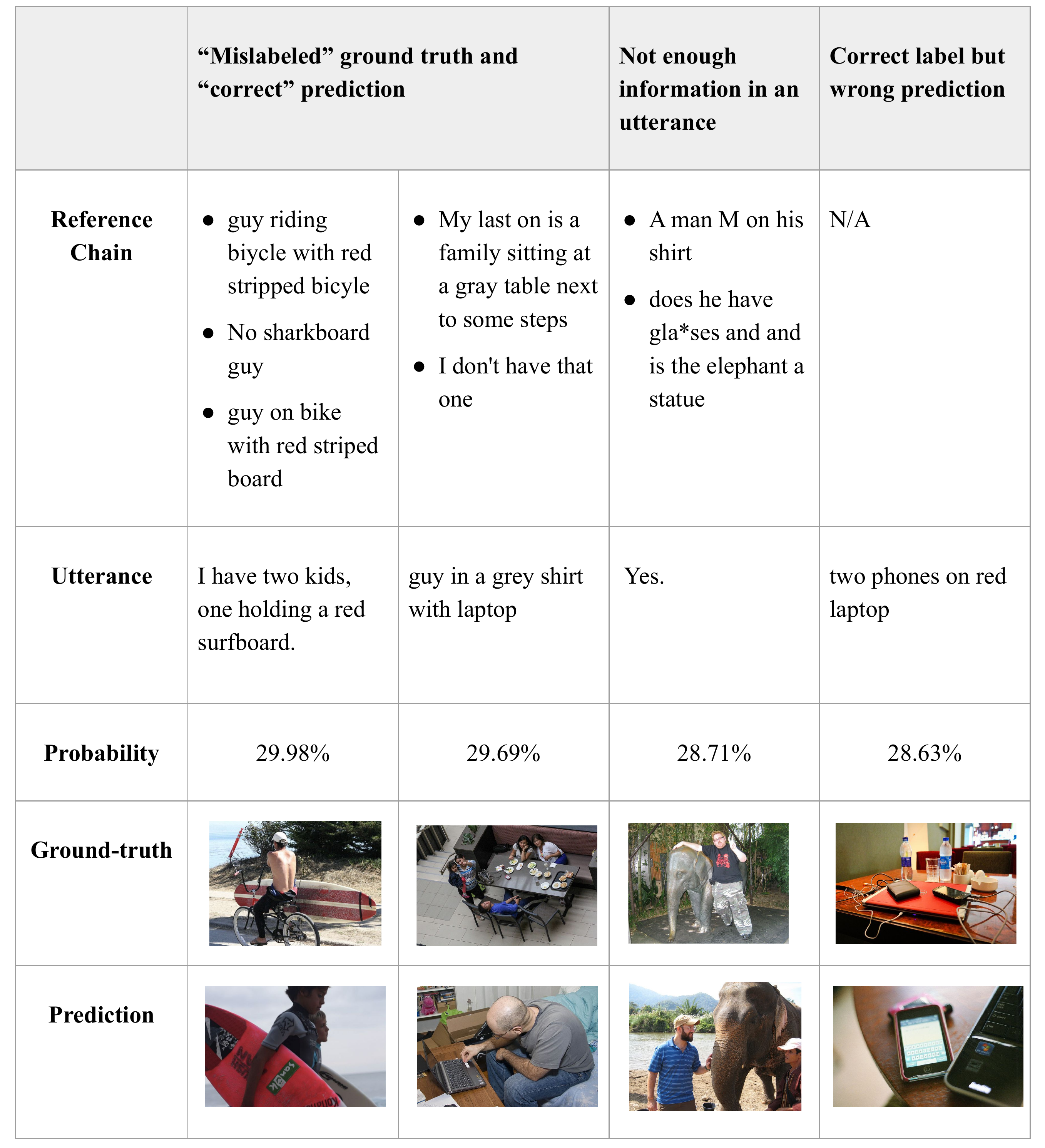}
    \caption{Examples of wrong predictions with the highest confidence by the reference resolution model proposed in \cite{takmaz2020refer}.
    Most of these cases can be attributed to errors arising from reference chain extraction.
    The `Utterance' input is actually the latest utterance in an extracted reference chain.
    ‘Probability’ means the probability assigned to the predicted image. }
    \label{tab:wrong_pred}
\end{table*}

\begin{table*}
\small
\setlength{\tabcolsep}{0.6em}
\centering
     \begin{tabular}{ l p{4.2cm}  p{4.2cm}  }
     \toprule
      Context and Target Images & Utterances & Labels and Predictions \\ \cmidrule(r){1-1}\cmidrule(lr){2-2}\cmidrule(l){3-3}
     \raisebox{-\totalheight}{\includegraphics[width=0.3\textwidth]{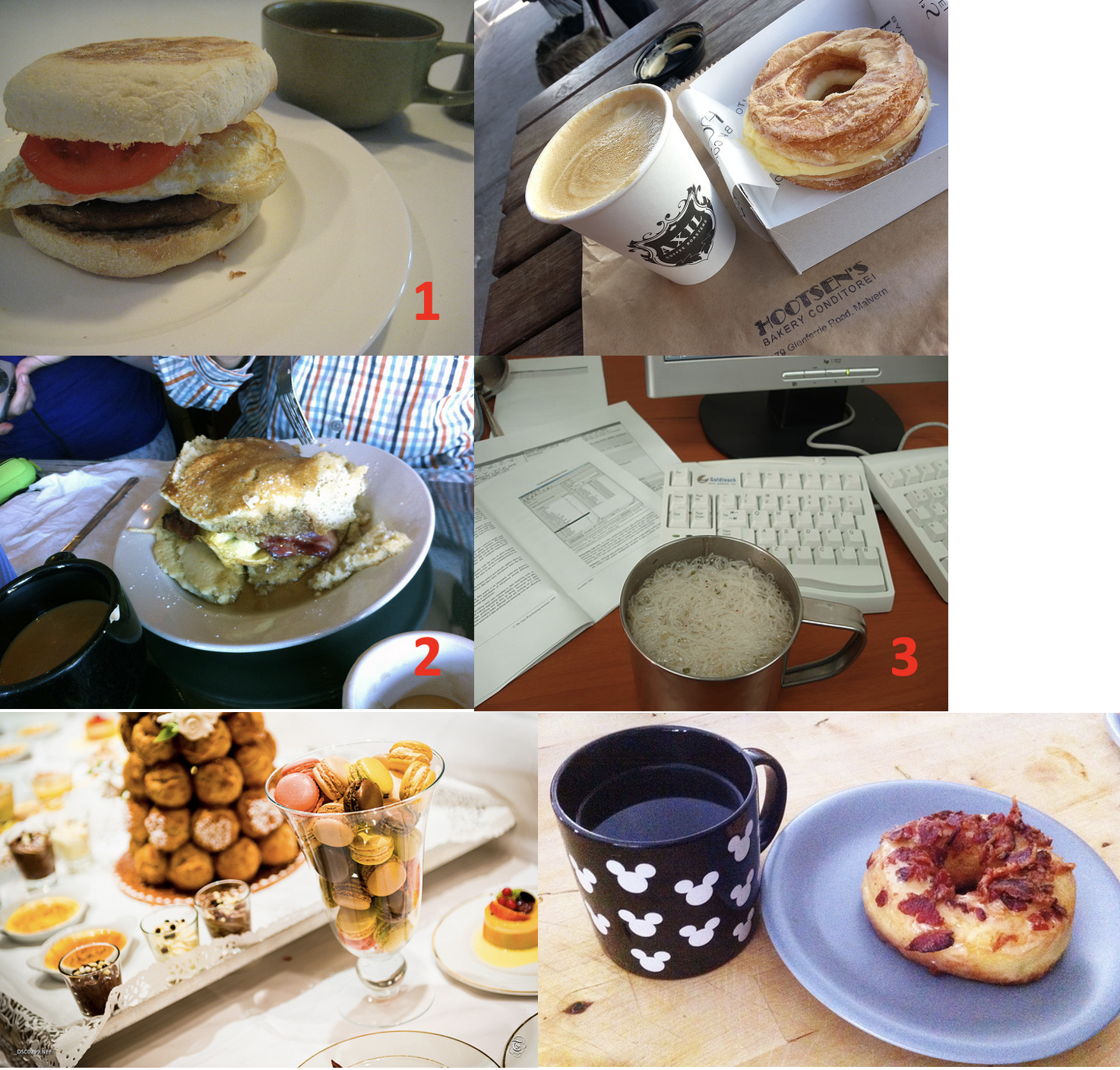}}
      &
      \begin{itemize}[nolistsep]
          \item A: I need to get my eyes checked lol
	   \item A: Okay so same english m**fin sandwhich
	   \item A: green tea mug
	   \item B: Nope
	   \item A: okay and I have the half keyboard latte one
	   \item B: yes
	   \item A: and the last one.. idk
	   \item A: it's a sandwhich but it looks like a mess
	   \item A: there is a black mug in the bottom left corner
	   \item B: Yup and something blue to the top left and striped to the top right? I have that
	   \item A: yeah that's it
	   \item A: that's all I have
	   \item B: Do you have the donut, with the blue mug and red/white staw?
	   \item A: nope
	   \item B: All done here too!
      \end{itemize}
      &
      True labels:
      \begin{itemize}[topsep=1pt, itemsep=0.5pt]
      \item Tgt.~Image 1: Different
      \item Tgt.~Image 2: Common
      \item Tgt.~Image 3: Common
      \end{itemize}
      \vspace{6mm}
      Model predictions:
      \begin{itemize}[topsep=1pt, itemsep=0.5pt]
      \item Tgt.~Image 1: Common
      \item Tgt.~Image 2: Different
      \item Tgt.~Image 3: Different
      \end{itemize} \\
      \cmidrule(r){1-1}\cmidrule(lr){2-2}\cmidrule(l){3-3}
     \raisebox{-\totalheight}{\includegraphics[width=0.3\textwidth]{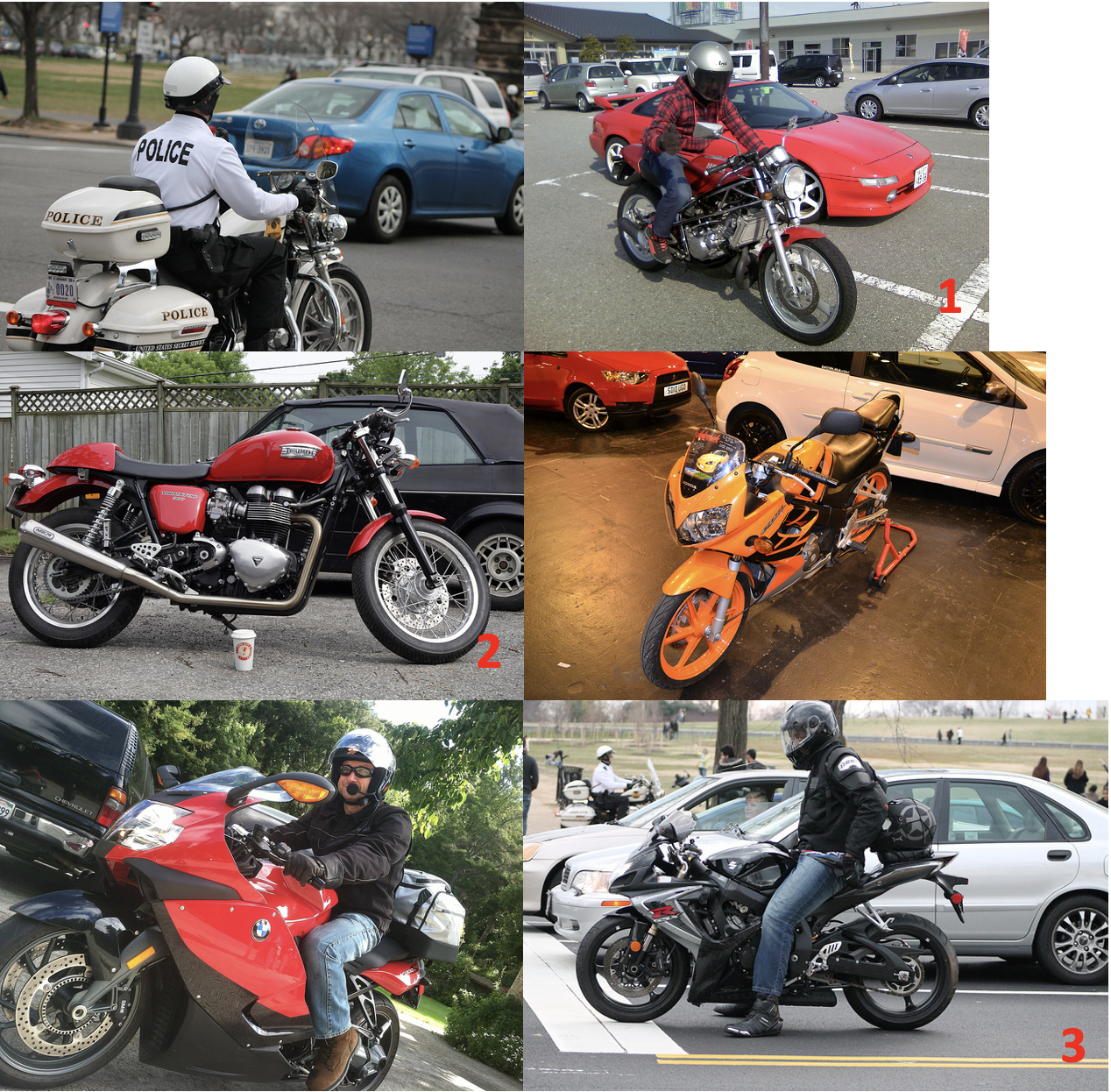}}
      &
      \begin{itemize}[nolistsep]
          \item B: I have the checkered shirt guy. do you have him?
	   \item A: do you have man vlack jacket and helmet next to silver car ?
	   \item A: Yes i do have the checkered shirt 
	   \item B: Is that the one at a gas station
	   \item A: no its on a street
	   \item B: oh then I don't have it
	   \item A: do you have red parked motorcycle in fornt of black car ?
	   \item B: Do you have one with a guy on a motorcycle in front of a gas station?
	   \item B: Yeah I have that one
	   \item A: no i do not have gas station
	   \item B: ok I'm set
	   \item A: me too
      \end{itemize}
      &
      True labels:
      \begin{itemize}[topsep=1pt, itemsep=0.5pt]
      \item Tgt.~Image 1: Common
      \item Tgt.~Image 2: Common
      \item Tgt.~Image 3: Different
      \end{itemize}
      \vspace{6mm}
      Model predictions:
      \begin{itemize}[topsep=1pt, itemsep=0.5pt]
      \item Tgt.~Image 1: Different
      \item Tgt.~Image 2: Different
      \item Tgt.~Image 3: Common
      \end{itemize}
      \\ \bottomrule
      \end{tabular}
      \caption{Selected examples on which our best-performing listener model predicts all 3 target images wrong. Both examples are from player A's view.
      Indices of target images are marked in {\color{red} red} in the image's lower-right corner.
      }\label{tbl:look_at_all_wrongs}
\end{table*}



\end{document}